\documentclass{article}

\usepackage[final]{neurips_2024}
\usepackage{amsmath}
\usepackage{graphicx}
\usepackage{multirow}
\usepackage{algorithm}
\usepackage{algorithmic}
\usepackage{graphicx}
\usepackage{caption}
\usepackage{subcaption}
\usepackage{float}  
\usepackage{amssymb}
\usepackage{hyperref}
\usepackage{cleveref}

\usepackage{xcolor}
\definecolor{mydarkblue}{rgb}{0,0.08,0.45}
\newcommand{\takeaway}[1]{\vspace{1mm}{\color[HTML]{1d5c38}\textbf{{$\triangleright$\hspace{5pt}#1}}}}

\title{Evolution Strategies for Deep RL
pretraining}

\author{
  Adrian Martínez, Ananya Gupta, Hanka Goralija, Mario Rico, \\
  Saúl Fenollosa, Tamar Alphaidze \\
  École Polytechnique Fédérale de Lausanne (EPFL)
}

\begin{document}

\maketitle
 \begin{abstract}

Although Deep Reinforcement Learning has proven highly effective for complex decision-making problems, it demands significant computational resources and careful parameter adjustment in order to develop successful strategies. Evolution strategies offer a more straightforward, derivative-free approach that is less computationally costly and simpler to deploy. However, ES generally do not match the performance levels achieved by DRL, which calls into question their suitability for more demanding scenarios.
This study examines the performance of ES and DRL across tasks of varying difficulty, including Flappy Bird, Breakout and Mujoco environments, as well as whether ES could be used for initial training to enhance DRL algorithms. The results indicate that ES do not consistently train faster than DRL. When used as a preliminary training step, they only provide benefits in less complex environments (Flappy Bird) and show minimal or no improvement in training efficiency or stability across different parameter settings when applied to more sophisticated tasks (Breakout and MuJoCo Walker).
The code is on \href{https://github.com/talphaidze/Evolution-Strategies-for-Deep-RL-Pretraining}{GitHub}.
 \end{abstract}

\vspace{-10pt}
\section{Introduction}
\vspace{-5pt}
Deep Reinforcement Learning (DRL) has shown remarkable success in tackling complex sequential decision-making problems \citep{sutton2018reinforcement}. However, this power comes with a cost, as it often requires extensive training and careful tuning to converge to effective policies~\citep{dulacarnold2019challengesrealworldreinforcementlearning, ghosh2021generalizationrldifficultepistemic}. In contrast, Evolution Strategies (ES) have surfaced as a simpler, gradient-free alternative that is computationally cheaper and easier to implement~\citep{salimans2017evolution}. These algorithms estimate gradients by performing a search over small parameter perturbations, thus avoiding the need to perform expensive gradient computations. However, ES methods typically fall short of the performance achieved by DRL, raising questions about their applicability to more challenging environments.

In this work, we evaluate and compare the performance of these two distinct paradigms: \textbf{DRL}, with algorithms such as Deep Q-Networks (DQN)~\citep{mnih2013playingatarideepreinforcement} and Proximal Policy Optimization (PPO)~\citep{schulman2017proximalpolicyoptimizationalgorithms}, and \textbf{ES}. We perform this analysis across environments that differ in complexity and reward structure. Our experiments span two main domains: arcade-style games with discrete action spaces, such as Flappy Bird and Breakout, and continuous control tasks in high-dimensional state and action spaces, modeled using the Mujoco physics simulator \citep{todorov2012mujoco}. Additionally, we investigate the applicability of ES as a pretraining strategy for DRL algorithms, in order to increase training speed and robustness to hyperparameter selection. 

Our findings show that ES is not consistently faster than DRL approaches, denying the first claim about training speed enhancement. Additionally, pretraining strategies appear to be effective for simpler environments, while they show little to no benefit in more complex tasks. In simple environments such as Flappy Bird, pretraining with ES appears to accelerate the learning curve for DRL algorithms, reaching higher rewards faster than DRL-only approaches. Unfortunately, this behaviour is not consistent on more complex environments, such as Breakout or Mujoco, where ES pretraining does not accelerate training and does not either make DRL algorithms more robust to hyperparameter selection.



\vspace{-5pt}
\section{Related Work}
\vspace{-5pt}
\subsection{Evolution Strategies}
\label{sec: ES}
Traditional reinforcement learning approaches typically optimize policy parameters using gradient-based methods, such as policy gradients or actor-critic algorithms \citep{sutton2018reinforcement}. These methods assume that the environment is smooth and differentiable with respect to actions, which allows gradients to propagate through trajectories using the chain rule. However, in many practical scenarios, such as physical simulations or real-world robotic systems, the environment may be non-differentiable, highly stochastic, or possess sparse and delayed rewards. In such cases, standard RL algorithms struggle to find effective policies \citep{dulac2019challenges}.

An alternative to gradient-based RL is to frame the learning problem as a black-box optimization task. In this setting, the policy is viewed as a mapping from states to actions, parameterized by $\theta \in \mathbb{R}^d$, and the objective is to maximize the expected cumulative reward obtained by executing the policy in the environment. Evolution Strategies (ES) offer a solution by searching for optimal parameters through random perturbations and selection based on observed rewards, without requiring access to environmental gradients or backpropagation \citep{salimans2017evolution}.

\subsubsection{Mathematical Formulation}

Let $F(\theta)$ denote the total reward achieved by running a policy with parameters $\theta$ in the environment. The goal is to find parameters $\theta$ that maximize $F(\theta)$. However, since $F$ is assumed to be a black-box function, meaning we do not have access to how the reward depends on individual actions or parameters, direct gradient computation is infeasible. To overcome this, Evolution Strategies optimize a smoothed version of the objective:

\begin{equation}
\tilde{J}(\theta) = \mathbb{E}_{\epsilon \sim \mathcal{N}(0, I)}[F(\theta + \sigma \epsilon)],
\end{equation}

where $\sigma > 0$ is a small noise parameter and $\epsilon \sim \mathcal{N}(0, I)$ is a standard multivariate Gaussian random vector. This formulation introduces smoothing by averaging over nearby parameter vectors, thereby making the objective amenable to gradient estimation.

The gradient of $\tilde{J}$ with respect to $\theta$ can be derived as follows. By the linearity of expectation and the chain rule:

\begin{equation}
\nabla_\theta \tilde{J}(\theta) = \mathbb{E}_{\epsilon}\left[ \nabla_\theta F(\theta + \sigma \epsilon) \right].
\end{equation}

Since $F$ is a black-box, we cannot directly compute $\nabla_\theta F(\theta + \sigma \epsilon)$. Instead, using the log-likelihood trick, we rewrite the gradient in terms of the score function:

\begin{equation}
\nabla_\theta \tilde{J}(\theta) = \mathbb{E}_{\epsilon}\left[ F(\theta + \sigma \epsilon) \nabla_\theta \log p_\theta(\theta + \sigma \epsilon) \right],
\end{equation}

where $p_\theta(\cdot)$ is the probability density function of $\mathcal{N}(\theta, \sigma^2 I)$.

Since the Gaussian log-probability satisfies:

\begin{equation}
\log p_\theta(x) = -\frac{1}{2\sigma^2} \|x - \theta\|^2 + \text{const} \implies \nabla_\theta \log p_\theta(x) = \frac{x - \theta}{\sigma^2}.
\end{equation}

Substituting $x = \theta + \sigma \epsilon$ yields:

\begin{equation}
\nabla_\theta \log p_\theta(\theta + \sigma \epsilon) = \frac{\sigma \epsilon}{\sigma^2} = \frac{\epsilon}{\sigma}.
\end{equation}

Thus, the gradient becomes:

\begin{equation}
\nabla_\theta \tilde{J}(\theta) = \frac{1}{\sigma} \mathbb{E}_{\epsilon} \left[ F(\theta + \sigma \epsilon) \cdot \epsilon \right].
\end{equation}

In practice, this expectation is approximated with Monte Carlo sampling. Given $n$ independent samples $\epsilon_1, \dots, \epsilon_n$, the gradient estimate is:

\begin{equation}
\nabla_\theta \tilde{J}(\theta) \approx \frac{1}{n \sigma} \sum_{i=1}^n F(\theta + \sigma \epsilon_i) \cdot \epsilon_i.
\end{equation}

\subsubsection{Algorithm}

The practical implementation of Evolution Strategies consists of sampling perturbations, evaluating the corresponding perturbed policies in the environment, estimating the gradient based on the collected rewards, and updating the policy parameters accordingly. The complete algorithm can be described as follows:

\begin{algorithm}
\caption{Evolution Strategies Algorithm}
\label{alg:es}
\begin{algorithmic}[1]
\STATE \textbf{Input:} Initial parameters $\theta_0$, learning rate $\alpha$, noise standard deviation $\sigma$, population size $n$
\FOR{iteration $t = 0, 1, 2, \dots$}
    \STATE Broadcast $\theta_t$ to all workers
    \FOR{each worker $i = 1$ to $n$ in parallel}
        \STATE Sample perturbation $\epsilon_i \sim \mathcal{N}(0, I)$
        \STATE Compute perturbed parameters $\theta_i = \theta_t + \sigma \epsilon_i$
        \STATE Execute policy with $\theta_i$ and obtain reward $F_i$
    \ENDFOR
    \STATE Aggregate all $(F_i, \epsilon_i)$ pairs
    \STATE Estimate gradient: $g_t = \frac{1}{n\sigma} \sum_i F_i \epsilon_i$
    \STATE Update parameters: $\theta_{t+1} = \theta_t + \alpha g_t$
\ENDFOR
\end{algorithmic}
\end{algorithm}

Because the evaluations of perturbed policies are independent, Evolution Strategies are naturally suited for parallel computation. In practice, to reduce variance, rewards $F_i$ are often replaced by rank-transformed and mean-centered values, a technique known as fitness shaping.
The current parameters $\theta_t$ are broadcast once to all workers at the start of each iteration. Each worker independently samples noise vectors and computes rewards locally.

To minimize communication overhead, workers do not transmit full parameter vectors. Instead, they communicate only the random seeds used to generate $\epsilon_i$ and the corresponding scalar rewards $F_i$. Given a common random seed initialization, all workers can deterministically reconstruct the sampled perturbations. This design ensures that only $O(1)$ data per rollout is communicated, making the method scalable to very high-dimensional policy spaces.

Unlike policy gradient methods, which inject noise into the action space at every timestep, Evolution Strategies perturb the policy parameters once at the beginning of each episode. This leads to a gradient estimator whose variance is independent of the episode length, making ES particularly robust in long-horizon or sparse-reward environments. Additionally, since ES avoids backpropagation through time, its gradient computation is significantly lighter than that of policy gradient methods.
A drawback, however, is that ES requires complete episode rollouts to compute returns, so the overall update can be delayed if even a single episode within the population takes a long time to finish.

\vspace{-5pt}
\section{Methodology}
\vspace{-5pt}

We aim to evaluate and compare the performance of gradient-based deep reinforcement learning (DRL) algorithms with gradient-free evolution strategies (ES). Specifically, we investigate two hypotheses: (1) whether ES can achieve intermediate performance benchmarks (e.g., reaching 25\% of the optimal reward) faster than DRL algorithms, and (2) whether ES can serve as an effective pretraining method to improve DRL training speed and robustness.

We conduct experiments across three benchmark environments of varying complexity. Concretely, we focus on two main environment domains: two arcade-style games with discrete action spaces, \textbf{Flappy Bird} and \textbf{Breakout}, and one continuous control task modeled using the \textbf{Mujoco} physics simulator \citep{todorov2012mujoco}.

For discrete action environments, we compare the performance of Deep Q-Networks (DQN) and the basic ES implementation, described in Section \ref{sec: ES}. In the continuous Mujoco domain, we use Proximal Policy Optimization (PPO) as the representative DRL method, since DQN is not applicable in continuous settings.

\subsection*{Performance Comparison}



We train DRL and ES agents from scratch under identical conditions, comparing them based on final reward, training time, and robustness to hyperparameters, environment, and seed variation. To evaluate ES as a pretraining method, we initialize DRL networks with parameters from ES-trained agents and assess whether this improves convergence or stability. Full implementation details are in Section 4.

\vspace{-0.3cm}
\section{Experimental Setup}
In this section, we present the configuration details for the environments used in our experiments. Specifically, \cref{sec:experimental_setup_flappy_bird} outlines the setup for the Flappy Bird environment, \cref{sec:experimental_setup_breakout} covers the Breakout environment, and \cref{sec;experimental_setup_mujoco} details the configuration for the MuJoCo environment. 

\label{sec:experimental_setup}
\vspace{-0.15cm}
\subsection{Flappy Bird}
In \cref{sec:flappy_bird_environment_details}, we describe the configuration of the Flappy Bird environment used in our experiments, including the state representation and reward structure. In \cref{sec:flappy_bird_model_arch}, we present the model architectures explored and explain how they were integrated into the hybrid approach. The parameters used for this environment can be seen in \cref{sec:flappy_bird_training_params}.
\label{sec:experimental_setup_flappy_bird}
\vspace{-0.15cm}
\subsubsection{Environment Details}
\label{sec:flappy_bird_environment_details}
Experiments were conducted in FlappyBirdEnv, a custom environment built with the Python Learning Game Engine that simulates the dynamics of the Flappy Bird game. Frame skipping was set to 4, repeating each action for four frames with cumulative rewards. The agent received +0.1 reward per timestep survived and +1.0 for each pipe passed, while collisions or failures resulted in a -1.0 penalty. This reward structure encouraged maximizing survival and progression.
\vspace{-0.15cm}
\subsubsection{Model Architecture}
\label{sec:flappy_bird_model_arch}
Both DQN and ES agents were based on a fully connected multilayer perceptron policy with two hidden layers of 64 units each and Tanh activation functions. The output layer produced a vector of scores corresponding to each discrete action. For DQN, this represented Q-values used for greedy or $\epsilon$-greedy action selection. In ES, the same output was used to deterministically select actions via argmax, with the reward used to guide population updates. This shared architecture facilitated weight transfer when initializing DQN from a pretrained ES policy.
\vspace{-0.15cm}
\subsection{Breakout}
\label{sec:experimental_setup_breakout}
In \cref{sec:breakout_environment_details}, we detail the configuration of the Breakout environment used in our experiments, including both image-based and RAM-based setups. \Cref{sec:breakout_model_arch} outlines the model architectures selected for each input type and their compatibility with the hybrid training framework. The parameters used for the breakout environment can be seen in \cref{sec:breakout_training_parmas}.
\vspace{-0.15cm}
\subsubsection{Environment Details}
\label{sec:breakout_environment_details}
The experiments were conducted using the Atari Breakout environment, a classic arcade game where the agent controls a paddle to bounce a ball and break bricks. The goal is to clear all bricks without letting the ball fall. Breakout is a standard reinforcement learning benchmark due to its visual complexity, sparse rewards, and need for precise control.

The action space includes four discrete actions: NOOP, FIRE, MOVE LEFT, and MOVE RIGHT. The agent earns +1 for each brick broken. An episode ends when all lives are lost or the level is cleared. To evaluate different input modalities, we used two environment variants:

For \textbf{Image-based setup (ALE/Breakout-v5)}, The agent observes stacked grayscale frames (84×84×4), capturing both spatial and temporal dynamics. This high-dimensional input emphasizes the role of vision and sequence learning.

For \textbf{RAM-based setup (Breakout-ram-v4)}, The agent receives a 128-dimensional vector representing the game’s internal memory state, offering a compact but semantically rich input. This setup tests whether ES performs better in low-dimensional, non-visual spaces.

\vspace{-0.25cm}
\subsubsection{Model Architecture}
\label{sec:breakout_model_arch}
For \textbf{Image-based setup (ALE/Breakout-v5)}, A convolutional neural network (CNN) was used to process stacked grayscale frames (84×84×4), following the standard DQN design. This architecture, consisting of three convolutional layers and two fully connected layers, includes approximately 850,000 parameters. It is well-suited for extracting spatial and temporal features but is computationally intensive.

For \textbf{RAM-based setup (Breakout-ram-v4)}, A multilayer perceptron (MLP) with two hidden layers of 256 units each was used for the 128-dimensional RAM input. This simpler architecture has around 90,000 parameters, making it more efficient but less expressive for complex spatial reasoning.

Both models were implemented under a shared interface supporting parameter access, perturbation, and evaluation, enabling use with both DQN and Evolution Strategies.

\vspace{-0.15cm}
\subsection{Mujoco Environments}
\label{sec;experimental_setup_mujoco}
In \cref{sec:environment_details_mujoco} and \cref{sec:model_arch_mujoco}, we describe how the MuJoCo environments from the Brax library were configured, trained, and architecturally implemented for our experiments. These environments were selected due to their diversity in dynamics, reward structures, and control complexity, making them ideal testbeds for evaluating and comparing gradient-based (PPO) and gradient-free (ES) reinforcement learning methods under consistent and reproducible conditions. The parameters are detailed in \cref{sec:Training_parameters_mujoco}.
\vspace{-0.25cm}
\subsubsection{Environment Details}
\label{sec:environment_details_mujoco}
We evaluated agent performance on three MuJoCo benchmark environments from the Brax library: HalfCheetah, Hopper, and Walker2d. Each environment involves planar locomotion in 2D with continuous control over joint torques. HalfCheetah tasks a 6-actuator robot with maximizing forward velocity and does not terminate episodes early. Hopper is a single-legged agent with a 3-dimensional action space, terminating episodes upon falling. Walker2d, a bipedal version of Hopper with six joints, also ends episodes if the agent becomes unstable. All environments allow configuration of key parameters like control cost, forward reward weight, and initial state noise. These tasks vary in dynamics and reward structure, making them well-suited for comparing gradient-free and gradient-based reinforcement learning methods. Furthermore, BRAX environments are differentiable, making them a fair comparision of ES and RL. 
\vspace{-0.25cm}
\subsubsection{Model Architecture}
\label{sec:model_arch_mujoco}

For all MuJoCo environments, we used a fully connected neural network as the policy architecture. The network consisted of 4 hidden layers, each with 32 units and non-linear activation functions. This architecture was kept lightweight to facilitate efficient optimization and compatibility across algorithms.

When using ES as a pretraining step for PPO, this network was employed as the actor component. The weights obtained from ES training were transferred to initialize the PPO actor. For the critic network, however, we initialized a separate randomly seeded network, allowing PPO to learn the value function independently. This setup enabled seamless integration between ES and PPO while preserving the actor-critic structure essential for PPO’s learning dynamics.
\section{Results}
\subsection{Flappy Bird}
\begin{figure}[htbp]
\centering
  \includegraphics[width=0.825\textwidth]{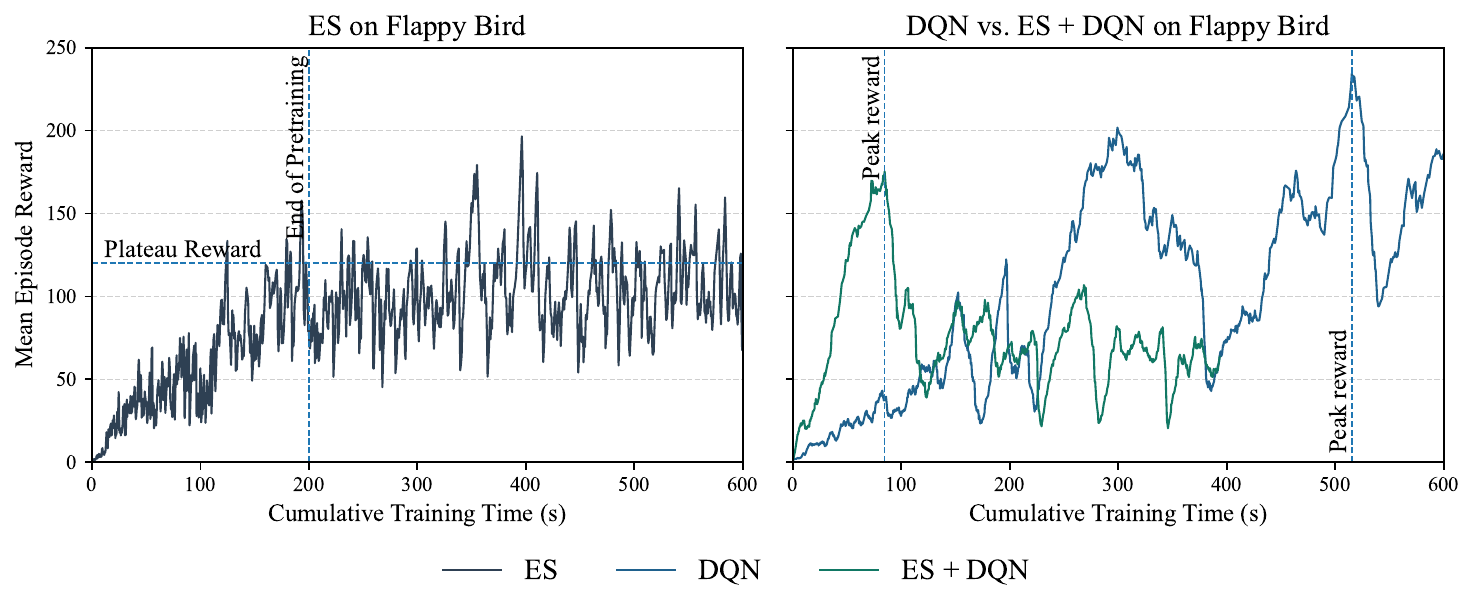}
  \caption{Smoothed learning curves for ES, DQN, and ES-pretrained DQN versus cumulative training time in Flappy Bird environment}
  \label{fig:flappy}
\end{figure}
In the Flappy Bird environment, as we can see on Figure \cref{fig:flappy}, ES demonstrated strong learning capabilities. Within a relatively short training time, ES was able to find a stable policy that achieved consistent survival and reward accumulation. While DQN eventually reached higher final rewards, it required significantly more training steps and showed greater sensitivity to hyperparameters and random seeds. When comparing the two methods, it’s important to note that DQN was trained in parallel across multiple environments, whereas ES was trained sequentially. Additionally, DQN experienced sudden drops in performance during training, from which it often did not recover. This kind of instability might be caused by large updates in the wrong direction, possibly due to overestimated Q-values or noisy gradients, which pushed the policy into worse regions and caused it to reinforce poor decisions through the replay buffer. When DQN was initialized with policy parameters obtained from ES, the agent reached competitive performance much faster compared to training from scratch. These results suggest that ES is particularly effective in simple, sparse-reward environments like Flappy Bird, where its stable and robust exploration helps provide a strong starting point for gradient-based fine-tuning.
\subsection{Breakout}


\begin{figure}[htbp]
    \centering
    \includegraphics[width=0.825\linewidth]{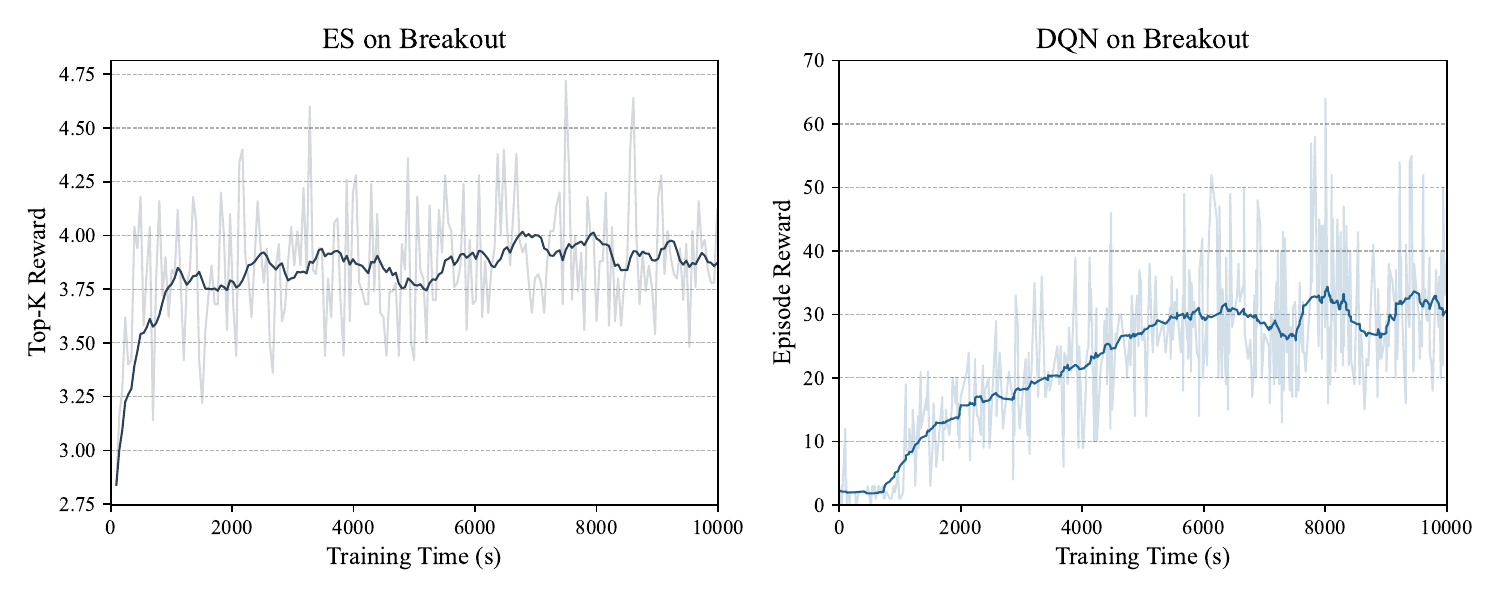}
    \caption{Smoothed learning curves for ES and DQN versus cumulative training time in Breakout environment}
    \label{fig:breakout}
\end{figure}


In our experiments, as shown in Figure \ref{fig:breakout}, the Deep Q-Network (DQN) with a CNN policy consistently achieved higher mean rewards of around 30, whereas Evolution Strategies (ES) plateaued at much lower rewards. From these observations, we highlight two main takeaways:

\takeaway{DQN consistently outperforms ES in Breakout, achieving mean rewards around 30 on the image-based environment.}
DQN’s CNN policy effectively handles high-dimensional pixel inputs, but its learning curve exhibits greater variance and is sensitive to hyperparameter tuning.

\takeaway{ES struggles to scale in complex settings, plateauing early in both pixel-based and RAM-based environments.}
ES applied to the CNN policy in the image-based environment plateaued at a mean reward near 1.5 despite a large population size. When using a simpler MLP policy on RAM-based inputs, ES showed faster initial progress with rewards reaching around 4 but failed to improve further. These results underscore ES’s difficulty extracting useful representations in high-dimensional and temporally extended settings, unlike the gradient-based updates and temporal difference learning used by DQN.



\vspace{-1.5mm}

\subsection{MuJoCo}



\begin{figure}[htbp]
    \centering

    \begin{subfigure}[t]{\linewidth}
        \centering
        \includegraphics[width=\linewidth]{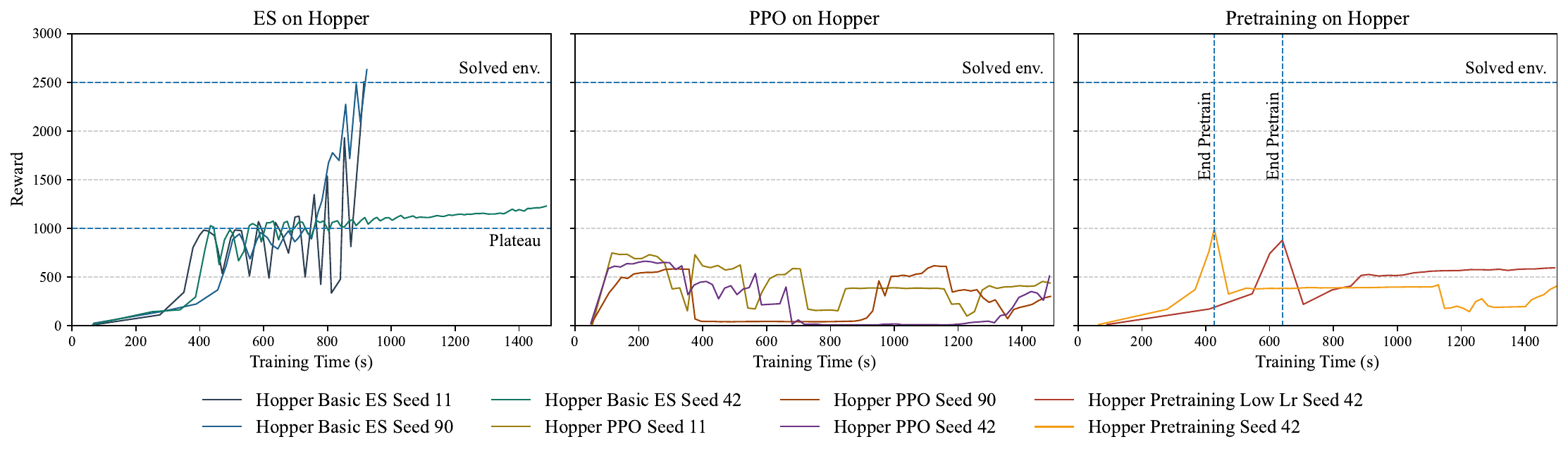}
        \caption{Smoothed learning curves for ES, PPO, and ES-pretrained PPO versus cumulative training time in Hopper environment.}
        \label{fig:hopper_es_ppo}
    \end{subfigure}

    \vspace{0.5em}

    \begin{subfigure}[t]{\linewidth}
        \centering
        \includegraphics[width=0.9\linewidth]{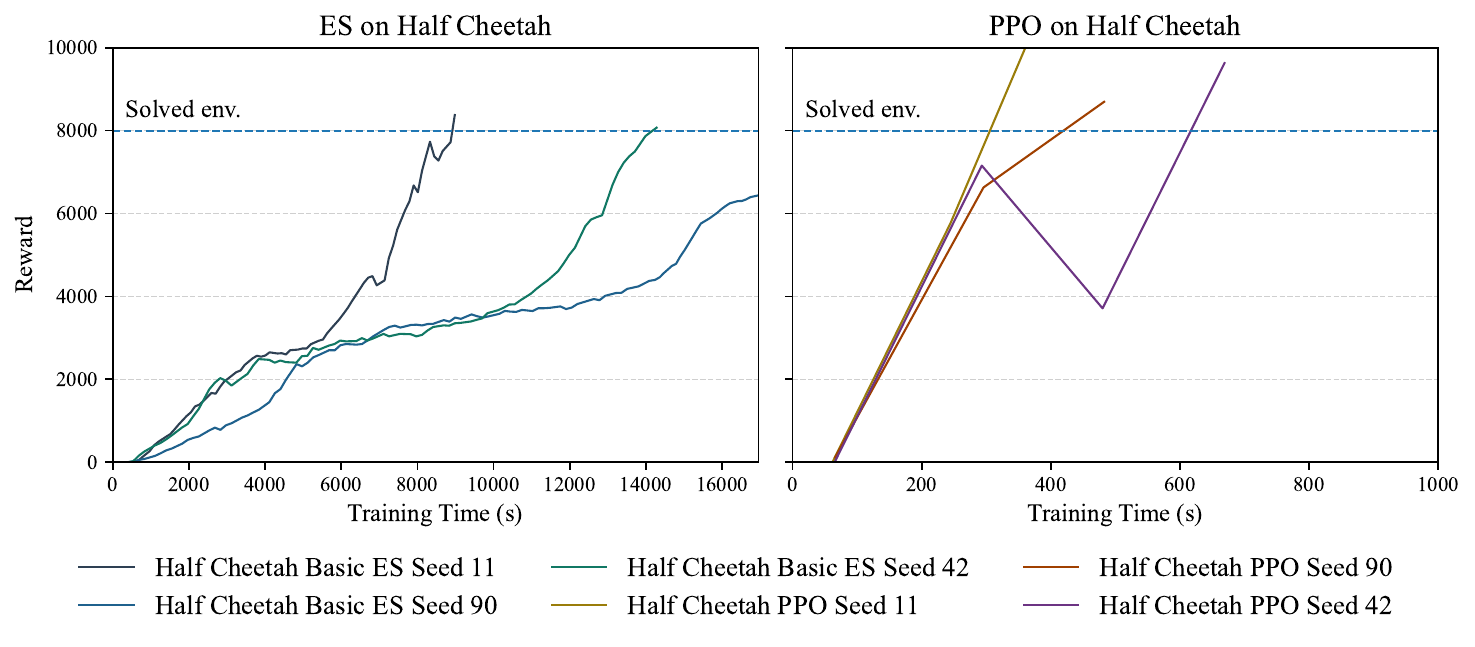}
        \caption{Smoothed learning curves for ES and PPO versus cumulative training time in Half Cheetah environment.}
        \label{fig:half_cheetah_es_ppo}
    \end{subfigure}

    \vspace{0.5em}

    \begin{subfigure}[t]{\linewidth}
        \centering
        \includegraphics[width=0.9\linewidth]{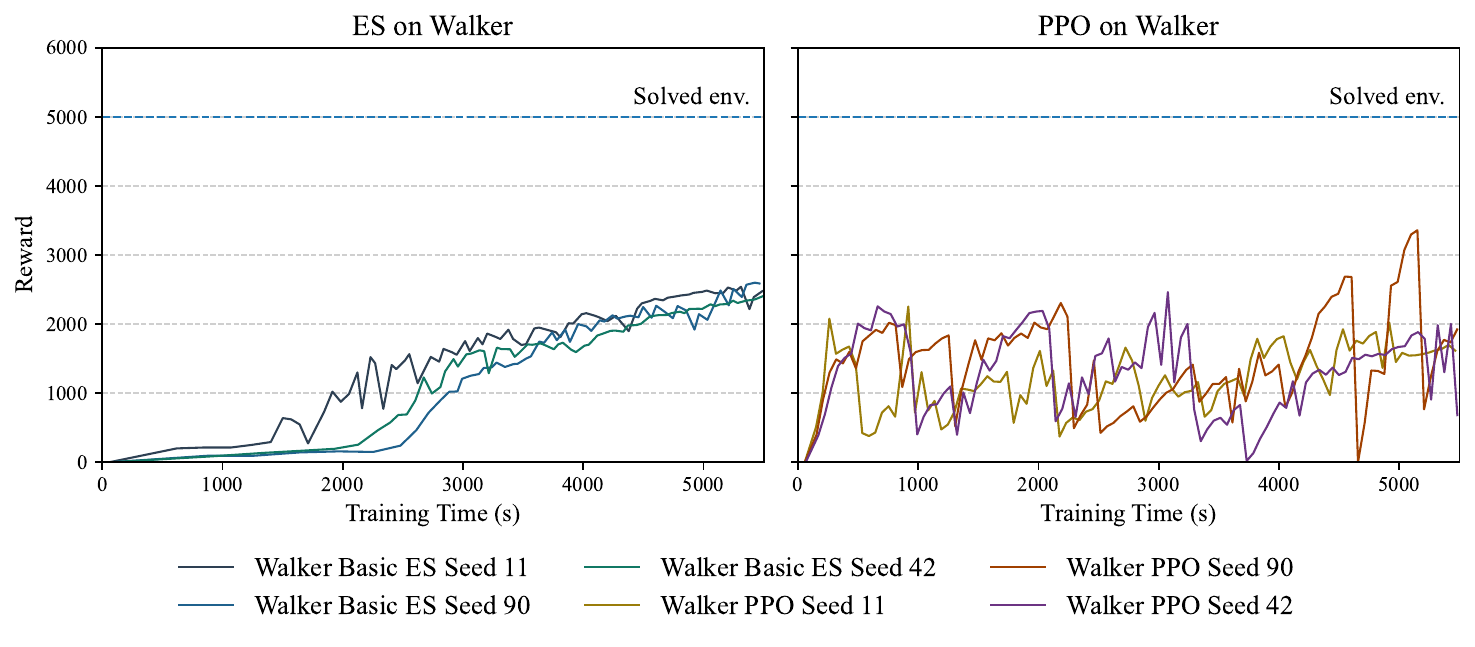}
                \caption{Smoothed learning curves for ES and PPO versus cumulative training time in Walker environment.}

        \label{fig:walker_es_ppo}
    \end{subfigure}

    \caption{Performance comparison of ES, PPO, and ES pretraining across various Mujoco environments.}
    \label{fig:mujoco_combined}
\end{figure}

Figures \ref{fig:mujoco_combined} showcases a performance comparison between PPO and ES on different Mujoco environments, described in Section \ref{sec;experimental_setup_mujoco}. After analyzing the obtained results, we come up with two main takeaways: 

\takeaway{PPO performs inconsistently across seeds and environments, while ES is slower but yields significantly more stable and repeatable outcomes.}

PPO demonstrates strong performance in some environments but can be unstable and highly sensitive to hyperparameter choices. In HalfCheetah (Figure \ref{fig:half_cheetah_es_ppo}), PPO converges 20x faster than ES. However, in other environments such as Walker2d (Figure \ref{fig:walker_es_ppo}) or Hopper (Figure \ref{fig:hopper_es_ppo}), PPO does not manage to converge and oscillates between low reward values. On the other hand, ES reliably solves most of the evaluated environments, failing to fully solve Walker2d. However, its convergence is much slower compared to PPO (20x slower in HalfCheetah).

\takeaway{Pretraining with ES does not improve PPO's training speed nor enhance robustness to hyperparameter selection.}  

While ES eventually solves most environments, it is significantly slower, up to 20× in HalfCheetah, and consistently fails to reach intermediate reward thresholds faster than PPO. This lag in early performance undermines its viability as a pretraining strategy for accelerating PPO's training. On the other hand, despite initializing PPO with parameters obtained from ES, the resulting training behavior remains largely unchanged (Figure \ref{fig:hopper_es_ppo}). PPO continues to exhibit similar sensitivity to hyperparameters, indicating that ES pretraining does not improve training stability. This results could stem from the fact that PPO uses an actor-critic, structure, while ES only accounts for the actor network optimization.

\noindent

\vspace{1em}

\vspace{-0.5em}

\vspace{-0.3cm}
\section{Limitations and next steps}
Although Evolution Strategies (ES) have shown strong performance in some DRL tasks - occasionally outperforming gradient-based methods - they were ineffective as a pretraining mechanism for RL, contrary to our hypothesis. We identify two main limitations:

First, ES and RL differ significantly in architecture and learning dynamics. In environments like Flappy Bird and MuJoCo, ES-trained policies did not transfer well due to differences in structure (e.g., PPO's separate value and policy networks vs. ES's single-policy optimization) and optimization methods (black-box search vs. gradient-based learning), leading to incompatible representations.
Second, ES performed poorly in complex environments such as Breakout. Even with vectorized (feature-based) inputs, results were weak, and performance degraded further with raw-pixel (CNN-based) inputs - highlighting ES’s difficulty in scaling to high-dimensional tasks.

Future work should explore adaptive, architecture-aware hybrid approaches that improve transfer between ES and RL, such as aligning learned representations, using shared modules, or modifying architectures to bridge the gap.

\vspace{-0.3cm}
\section{Conclusion}

Our results show that ES can effectively accelerate early learning in simple environments like Flappy Bird, especially when used to initialize gradient-based methods like DQN. However, ES struggled to scale to high-dimensional tasks like Breakout and was not compatible with PPO due to architectural differences. While DQN and PPO achieved better final performance overall, they were more sensitive to hyperparameters and showed variability across runs. These findings suggest that ES is a useful tool for exploration in low-complexity settings, but combining it with gradient-based methods in more complex tasks remains a challenge. Future work could explore adaptive or architecture-aware hybrid approaches that improve transfer between ES and deep RL algorithms.

\newpage
\bibliographystyle{chicago}
\bibliography{ref}

\newpage
\appendix
\section{Deep Q-Networks (DQN)}

The Deep Q-Network (DQN) algorithm combines Q-learning with deep neural networks to handle high-dimensional state spaces in reinforcement learning. It approximates the action-value function $Q(s, a; \theta)$ with a neural network parameterized by weights $\theta$, where $s$ is the current state and $a$ is the action taken.

\subsection{Mathematical Formulation}

The goal is to learn an optimal policy that maximizes the expected return by estimating the action-value function:

\begin{equation}
Q^*(s, a) = \max_{\pi} \mathbb{E} \left[ \sum_{t=0}^{\infty} \gamma^t r_t \,|\, s_0 = s, a_0 = a, \pi \right],
\end{equation}

where $\gamma \in [0, 1]$ is the discount factor and $r_t$ is the reward at timestep $t$. In Q-learning, the optimal $Q$-function satisfies the Bellman equation:

\begin{equation}
Q^*(s, a) = \mathbb{E}_{s'} \left[ r + \gamma \max_{a'} Q^*(s', a') \,|\, s, a \right].
\end{equation}

DQN minimizes the temporal-difference (TD) error between the current estimate $Q(s, a; \theta)$ and the target value $y = r + \gamma \max_{a'} Q(s', a'; \theta^{-})$, where $\theta^{-}$ are the parameters of a separate target network. The loss function is:

\begin{equation}
L(\theta) = \mathbb{E}_{(s,a,r,s') \sim D} \left[ \left( y - Q(s, a; \theta) \right)^2 \right],
\end{equation}

where $D$ is the experience replay buffer. The target $y$ is treated as a fixed value during the optimization step:

\begin{equation}
y = r + \gamma \max_{a'} Q(s', a'; \theta^{-}).
\end{equation}

To stabilize training, DQN introduces two key mechanisms:

\begin{itemize}
    \item \textbf{Experience Replay:} A buffer $D$ stores past transitions $(s, a, r, s')$, and mini-batches are sampled uniformly from it to break correlation between sequential data.
    \item \textbf{Target Network:} A separate network with parameters $\theta^{-}$ is used to compute the target values. Its parameters are periodically updated to match $\theta$.
\end{itemize}

\subsection{Algorithm}

The DQN algorithm can be summarized as follows:

\begin{algorithm}
\caption{Deep Q-Network (DQN)}
\label{alg:dqn}
\begin{algorithmic}[1]
\STATE Initialize replay buffer $D$ with capacity $N$
\STATE Initialize Q-network with weights $\theta$
\STATE Initialize target Q-network with weights $\theta^{-} = \theta$
\FOR{each episode}
    \STATE Initialize environment and receive initial state $s$
    \FOR{each timestep}
        \STATE With probability $\epsilon$ select a random action $a$, otherwise $a = \arg\max_{a} Q(s, a; \theta)$
        \STATE Execute action $a$, observe reward $r$ and next state $s'$
        \STATE Store transition $(s, a, r, s')$ in $D$
        \STATE Sample random minibatch from $D$
        \STATE Compute target $y = r + \gamma \max_{a'} Q(s', a'; \theta^{-})$
        \STATE Perform gradient descent on $\left( y - Q(s, a; \theta) \right)^2$
        \STATE Update $s \leftarrow s'$
        \STATE Every $C$ steps: $\theta^{-} \leftarrow \theta$
    \ENDFOR
\ENDFOR
\end{algorithmic}
\end{algorithm}

By combining off-policy Q-learning with deep neural networks, DQN has been successfully applied to challenging domains such as Atari games from raw pixels, achieving human-level performance in many cases.

\section{Parameters}
\label{sec:Parameters}
In \cref{sec:flappy_bird_training_params}, we detailed the hyperparameters used to train agents in the Flappy Bird environment. \Cref{sec:breakout_training_parmas} presents the training parameters for the Breakout experiments, covering both DQN and ES configurations. Finally, \cref{sec:Training_parameters_mujoco} outlines the training setup adopted for the MuJoCo environments, including environment-specific hyperparameters and standardized PPO configurations from the Brax benchmark suite. Together, these sections provide a comprehensive overview of the experimental conditions under which each algorithm was evaluated.

\subsection{Flappy Bird Training Parameters}
\label{sec:flappy_bird_training_params}
The Deep Q-Network (DQN) agent was trained for 1,000,000 timesteps, corresponding to approximately 5000 episodes. Training was parallelized across 8 synchronous environments to speed up sample collection and stabilize learning. The agent followed an $\epsilon$-greedy exploration strategy, with $\epsilon$ annealed linearly from 0.2 to 0.0001 over the first 10\% of training. Additional hyperparameters included a learning rate of 5e-5, a replay buffer size of 10,000 transitions, a mini-batch size of 32, and a discount factor $\gamma$ of 0.90.

The Evolution Strategies (ES) algorithm used a population size of 16 and a Gaussian noise standard deviation of $\sigma=0.05$. It was trained sequentially for 1000 generations with a learning rate of 0.005. ES operated directly on the policy parameters, using returns as a fitness measure to guide the search. After training, the trained ES policy was used to initialize the hidden layers of the DQN network. This provided a warm start for DQN, allowing it to begin learning from a reasonably good policy rather than from scratch.
\subsection{Breakout Training Parameters}
\label{sec:breakout_training_parmas}
For DQN, we trained the agent using the following hyperparameters: a learning rate of 2e-4, buffer size of 100,000 transitions, batch size of 32, training frequency of every 4 steps, and one gradient update per training step. Target network updates were performed every 1,000 steps. Training followed an $\varepsilon$-greedy policy, where $\varepsilon$ decayed from 0.1 to 0.01 over time. DQN training was managed through Stable-Baselines3 and included support for both CnnPolicy and MlpPolicy depending on the input type.

For Evolution Strategy, training was conducted over 500 generations with a population size of 50. Each individual was evaluated over 5 episodes, and symmetric Gaussian noise was applied to the model parameters using a standard deviation (sigma) of 0.2. The learning rate for ES updates was set to 0.01. Parameters were updated using the reward-weighted average of perturbations, normalized by the population standard deviation. Checkpoints were saved every 100 generations, and metrics were logged using Weights \& Biases (wandb). For both ES and DQN we tried different hyperparameters and arrived on the parameters specified in this report.

\subsection{Mujoco Training Parameters}
\label{sec:Training_parameters_mujoco}

For all MuJoCo tasks, we followed the default PPO training configurations provided in the Brax benchmark suite \citep{freeman2021braxdifferentiablephysics}. These setups were optimized for efficient and stable learning in continuous control environments using large-scale parallel simulation (num envs = 8192).
\begin{itemize}
\item \textbf{Walker2d}: Trained for $7{,}864{,}320$ timesteps with $20$ evaluation points. The reward was scaled by $5$, and the discount factor was $\gamma = 0.997$. We used a learning rate of $6 \times 10^{-4}$, batch size $128$, and $32$ gradient updates per environment step. Training began after $8192$ transitions were collected, with a replay buffer of size $2^{20}$.

\item \textbf{HalfCheetah} and \textbf{Hopper}: Both trained for $6{,}553{,}600$ timesteps using similar settings, but with a higher reward scaling factor of $30$, batch size $512$, and $64$ gradient updates per step to support faster locomotion dynamics.
\end{itemize}

All environments used observation normalization to stabilize training, with each state input standardized to reduce variance across features. Actions were applied at every simulation step without repetition (i.e., action repeat was set to 1), ensuring fine-grained control. A fixed random seed (set to 1) was used to ensure reproducibility of results. Training was constrained to one computational device per host, matching the single-device setup used in the original Brax benchmarks. These standardized settings ensured fair comparison between the PPO and ES approaches across all environments.

\end{document}